\documentclass[9pt,conference]{IEEEtran}
\IEEEoverridecommandlockouts
\usepackage{graphicx}
\usepackage{amssymb}
\usepackage{multirow}
\usepackage{array}
\usepackage{indentfirst}
\usepackage{svg}
\usepackage{float}
\usepackage[colorlinks,citecolor=green,urlcolor=blue,bookmarks=false,hypertexnames=true]{hyperref} 
\newcommand{\Real}{{\rm I\!R}}
\usepackage{amsmath}

\usepackage[hang,flushmargin]{footmisc}

\usepackage{authblk}
\usepackage{setspace}
\usepackage[normalem]{ulem}
\useunder{\uline}{\ul}{}

\begin{document}

\title{CheapNVS: Real-Time On-Device Narrow-Baseline Novel View Synthesis}
\author[$\dagger$ $^*$]{Konstantinos Georgiadis}
\author[$\ddagger$ $^*$]{Mehmet Kerim Yucel}
\author[$\ddagger$ $^*$]{\vspace{-3.5mm}Albert Sa\`a-Garriga}
\affil[$\dagger$]{CERTH, Information Technologies Institute, Thessaloniki, Greece}
\affil[$\ddagger$]{Samsung R\&D Institute UK (SRUK) \vspace{-0.5mm}}
\affil[$^*$]{\small Three authors contributed equally.}

\maketitle
\begin{abstract}
Single-view novel view synthesis (NVS) is a notorious problem due to its ill-posed nature, and often requires large, computationally expensive approaches to produce tangible results. In this paper, we propose \textbf{CheapNVS}: a fully end-to-end approach for narrow baseline single-view NVS based on a novel, efficient multiple encoder/decoder design trained in a multi-stage fashion. CheapNVS first approximates the laborious 3D image warping with lightweight learnable modules that are conditioned on the camera pose embeddings of the target view, and then performs inpainting on the occluded regions in parallel to achieve significant performance gains. Once trained on a subset of Open Images dataset, CheapNVS outperforms the state-of-the-art despite being 10 $\times$ faster and consuming 6\% less memory. Furthermore, CheapNVS runs comfortably in real-time on mobile devices, reaching over 30 FPS on a Samsung Tab 9+.  
\end{abstract}

\begin{IEEEkeywords}
Novel View Synthesis; 3D Photography \footnote{\tiny 2025 IEEE.  Personal use of this material is permitted.  Permission from IEEE must be obtained for all other uses, in any current or future media, including reprinting/republishing this material for advertising or promotional purposes, creating new collective works, for resale or redistribution to servers or lists, or reuse of any copyrighted component of this work in other works.}
\end{IEEEkeywords}

\vspace{-2mm}
\section{Introduction} \label{introduction}
\noindent The ability to synthesize novel views of a scene/object from a single image is critical in computer vision, with applications in robotics \cite{adamkiewicz2022vision}, VR \cite{li2023instant}, multimedia production \cite{skartados2024finding} and biomedicine \cite{corona2022mednerf}. Novel view synthesis (NVS) is an ill-posed problem, as it is not only comprised of reverse-projecting an image onto the 3D space, but also of the completion of the missing data in occluded regions. Despite the recent advances in NVS \cite{mildenhall2021nerf, kerbl20233d, fan2023pose}, there are several issues with existing methods: they either i) require per-scene training and lack generalization across-scenes \cite{mildenhall2021nerf,kerbl20233d}, ii) overfit to fixed stereo baselines \cite{zhou2023single,chen2022pseudo,evain2019lightweight,chaurasia2020passthrough}, iii) are unfit for on-device, real-time processing \cite{mildenhall2021nerf, kerbl20233d, fan2023pose,zhou2023single,shih20203d,han2022single,shi2023zero123,tang2024mvdiffusion}  or iv) are not end-to-end learnable solutions \cite{zhou2023single,kopf2020one,chaurasia2020passthrough}.

\begin{figure*}[!ht]
  \centering
    \includegraphics[width=0.95\textwidth]{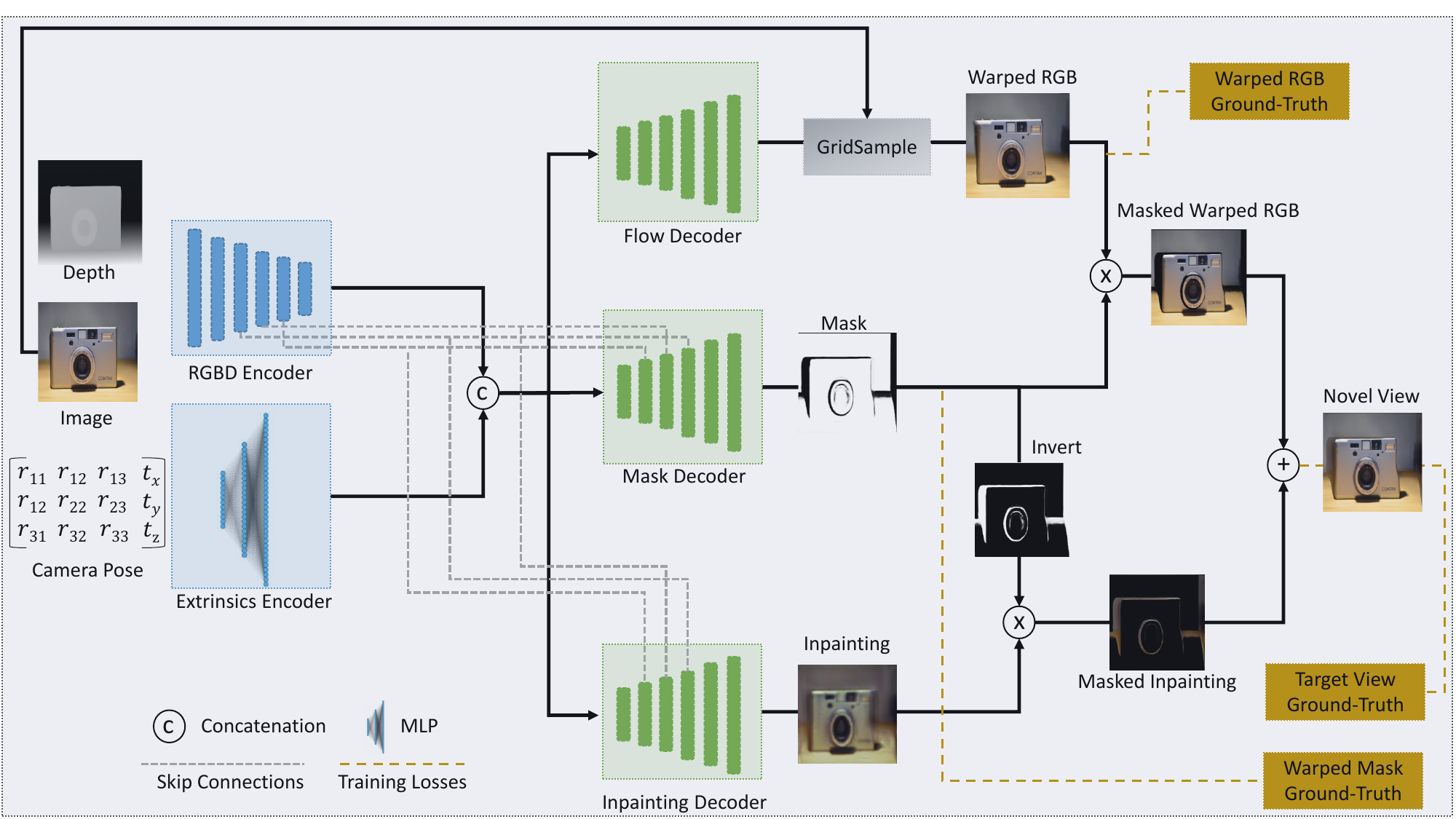}
  \vspace{-3mm}
  \caption{Our CheapNVS architecture. CheapNVS embeds target camera pose and RGBD input information into a shared latent space, which is then used by flow and mask decoders to perform learnable warping, and inpainting decoder to fill in occluded areas. Performing inpainting and warping in parallel, as well as approximating 3D warping via flow and mask decoders make CheapNVS more computationally friendly than existing methods.}
  \vspace{-4mm}
  \label{fig:overall_diagram}
\end{figure*}

In this paper, we aim to take a step towards meeting all the criteria above. To this end, we propose CheapNVS; an end-to-end, device-friendly, real-time solution for narrow baseline NVS. First, CheapNVS introduces an efficient warping module that approximates 3D image warping, conditioned on the camera pose of the target view, which speeds up the NVS pipeline. Second, unlike the contemporary methods that perform inpainting after the warping \cite{kopf2020one,han2022single,shih20203d}, CheapNVS performs inpainting in parallel, enjoying further speed-ups. Third, we propose to use Open Images as our training set. Based on a novel architecture that consists of a shared RGBD encoder, an extrinsics encoder, and three decoders, CheapNVS enjoys significant runtime and memory improvements, with a competitive or better accuracy to state-of-the-art. It runs comfortably in real-time on mobile devices, and since it is built on simple network blocks, can leverage hardware accelerators without requiring any low-level optimization expertise. The diagram of CheapNVS is shown in Figure \ref{fig:overall_diagram}.

\vspace{-1mm}

\section{Related Work} \label{related_work}

\noindent \textbf{Novel View Synthesis.} NVS is arguably the ultimate aim of 3D reconstruction, where a plethora of methods are proposed, ranging from earlier methods \cite{schoenberger2016sfm} to neural renderers \cite{mildenhall2021nerf,kerbl20233d}. Narrow baseline NVS, where the target view is not \textit{distant} from the source view, does not necessarily require exhaustive 3D reconstruction. This has been exploited in many studies, including multi-view NVS \cite{jantet2009incremental,flynn2019deepview} and 3D photography \cite{kopf2020one,shih20203d,han2022single}, to achieve performant methods.

\noindent \textbf{Single-view Novel View Synthesis.} Single-view NVS aims to do the same but using a single image. Stereo synthesis methods \cite{zhou2023single,chen2022pseudo,evain2019lightweight,chaurasia2020passthrough} perform NVS for a fixed baseline - mostly including only translation - so they only work for the baseline they are trained on. Modern one-shot neural rendering methods based on diffusion models provide good results, but they are computationally expensive or require per-scene optimization, thus not suitable for device deployment \cite{li2023instant,tang2024mvdiffusion}. 

The most relevant literature to ours is 3D photography \cite{kopf2020one,shih20203d,han2022single}. Our work differs from these in the following aspects; CheapNVS i) efficiently approximates 3D warping, whereas others perform 3D warping in the conventional, expensive way, ii) performs inpainting in parallel, not sequentially like others and iii) does away with representations such as Multi-Plane Images (MPI) \cite{han2022single} and Layered Depth Images (LDI) \cite{kopf2020one}, yet shows competitive results with significant efficiency gains and iv) comfortably runs in real-time on mobile devices, whereas others struggle to do so.

\vspace{-1mm}

\section{CheapNVS}
\label{methodology}

\subsection{Preliminaries}
\noindent Given an input image $\mathbb{I}_{s} \in \Real^{H \times W \times 3}$ and its depth map $\mathbb{D} \in \Real^{H \times W}$, single-view NVS aims to synthesize a new image $\mathbb{I}_{t} \in \Real^{H \times W \times 3}$ of the same scene from a given target camera pose $\mathbb{T} \in \Real^{3 \times 4}$. We can write single-view NVS as  
\begin{equation} \label{eq:nvs}
    \mathbb{I}_{t} = f(w(\mathbb{I}_{s}, \mathbb{T}, \mathbb{D}; \theta_{w}), \mathbb{M}; \theta_{f}) 
\end{equation}

where $w(\cdot;\theta_{w})$ is the function that implements image warping, $\mathbb{M} \in \Real^{H \times W}$ is the occlusion mask which indicates the areas to be \textit{filled} in the warped image, and $f(\cdot; \theta_{f})$ is the function that performs the filling to synthesize the novel view.

\subsection{A better NVS formulation.}

\noindent The primary issue of Equation \ref{eq:nvs} is that $f(\cdot; \theta_{f})$  requires the output of $w(\cdot;\theta_{w})$, which makes the process inherently sequential. We hypothesize that we can perform these in parallel, and have a modular and an interpretable architecture. An alternative to Equation \ref{eq:nvs} is 
\vspace{-1mm}
\begin{equation} \label{eq:cheapnvs}
\begin{aligned}
&\mathbb{I}_{t} = w(\mathbb{I}_{s}, \mathbb{T}, \mathbb{D}; \theta_{w}) \cdot \phi(\mathbb{I}_{s}, \mathbb{T}, \mathbb{D}; \theta_{\phi}) +  \\ 
&\quad \; \; \; \; f(\mathbb{I}_{s}, \mathbb{T}, \mathbb{D}; \theta_{f}) \cdot (1 - \phi(\mathbb{I}_{s}, \mathbb{T}, \mathbb{D}; \theta_{\phi}))  
\end{aligned}
\end{equation}


where $\phi(\cdot; \theta_{\phi})$ is the function that outputs the occlusion mask. Reminiscent of the image blending formulation \cite{georgiadis2022adaptive,yucel2023lra}, note that both $w(\cdot;\theta_{w})$, $f(\cdot; \theta_{f})$ and $\phi(\cdot; \theta_{\phi})$ take the same inputs, which shows the possibility of a shared backbone, which is better for both the knowledge sharing between all functions and computational savings. Furthermore, this formulation is modular, as one can obtain the occlusion mask, warped image and inpainted textures separately, unlike the recent diffusion-based NVS  solutions \cite{li2023instant,tang2024mvdiffusion}.

\subsection{CheapNVS}
\noindent CheapNVS performs single-view NVS by implementing the formulation of Equation \ref{eq:cheapnvs}. More specifically, we exploit the inherent parallelism of Equation \ref{eq:cheapnvs} and propose to implement
\vspace{-4mm}

\begin{equation}
\begin{split}
  \mathbb{M} = \phi(&\mathbf{F};\theta_{\phi})
  \\
  \mathbb{S} = \overline{w}(&\mathbf{F};\theta_{\overline{w}})
  \\
  \mathbb{\mathbb{P}} = f(&\mathbf{F};\theta_{f})\\
    \mathbf{F} = concat(\delta(\mathbb{I}_{s}, &\mathbb{D}; \theta_{\delta}), \nabla(\mathbb{T}; \theta_{\nabla}))  \\
\end{split}
\end{equation} \label{eq:cheapnvs_parts}
\vspace{-1mm}
where $\mathbb{S} \in \Real^{H \times W}$, $\mathbb{P} \in \Real^{H \times W}$, $\mathbf{F}$ and $\overline{w}(\cdot; \theta_{\overline{w}})$ are the predicted flow, inpainting, the shared embedding and the flow predictor, respectively. $\mathbf{F}$ is obtained by concatenating the outputs of $\delta(\cdot;\theta_{\delta})$ and $\nabla(\cdot;\theta_{\nabla})$, which encode input RGBD and the camera transformation matrix, respectively. We then synthesize the target view as 
\vspace{-1mm}
\begin{equation}
        \mathbb{I}_{t} = gs(\mathbb{S}, \mathbb{I}_{s}) \cdot \mathbb{M} + \mathbb{P} \cdot (1 - \mathbb{M})
\vspace{-0.6mm}
\end{equation} \label{eq:final_eq}
\vspace{-0.6mm}
where $gs(\cdot)$ is the grid sampling function (we use PyTorch in practice) that warps the image $\mathbb{I}_{s}$ via the shift map $\mathbb{S}$. We implement $\delta(\cdot)$, $\nabla(\cdot)$, $\overline{w}(\cdot)$, $f(\cdot)$ and $\phi(\cdot)$ with neural networks by learning their parameters $\theta_{\delta,\phi,\nabla,f,\overline{w}}$ in an end-to-end manner. 

\vspace{-2mm}
\subsection{The architecture}

\noindent \textbf{RGBD encoder.} RGBD encoder takes in an input image and the its corresponding depth map, and maps these into a shared latent space. This latent space is shared between all three decoders, which facilitates an implicit knowledge sharing between the decoders and also saves computational budget. We use a MobileNetv2 \cite{sandler2018mobilenetv2} to implement the RGBD encoder. 

\noindent \textbf{Extrinsics encoder.}
The image warping process is conditioned on the camera pose of the target view. Although the conventional warping methods use this information \cite{han2022single}, learnable warping methods discard this as they work for fixed baselines \cite{zhou2023single,chen2022pseudo,evain2019lightweight,chaurasia2020passthrough}. Our extrinsics encoder fixes that issue by learning an embedding for the target camera view. Specifically, it takes in the transformation matrix (between $\mathbb{I}_{s}$ and $\mathbb{I}_{t}$) and produces an embedding, which is then concatenated with the output of the RGBD encoder. We implement the extrinsics encoder with a cheap, two-layer MLP, which progressively maps the 12-D input to 256-D output.

\noindent \textbf{Discussion.} The two-encoder approach facilitates parallelism, and saves compute compared to having a single, large encoder for both RGBD and camera pose inputs. Additionally, we use skip connections from the RGBD encoder to mask and inpainting decoders to facilitate better knowledge sharing. We get negative returns when we have skip connections to the flow decoder, which we discuss in Section \ref{sec:ablations}.

\begin{table*}[!t]
\resizebox{\textwidth}{!}{%
\begin{tabular}{l|clllll|clllll|ccc}
& \multicolumn{6}{c|}{Open Images}                                                                                                                                                 & \multicolumn{6}{c|}{COCO}                                                                                                             & \multicolumn{3}{c}{Performance}                                      \\
& \multicolumn{3}{c}{Warping}                                                        & \multicolumn{3}{c|}{Inpainting}                                                             & \multicolumn{3}{c}{Warping}                                                        & \multicolumn{3}{c|}{Inpainting}                  & \multicolumn{2}{c}{Runtime}                 & Memory                 \\ \hline
Method                                                  & \multicolumn{1}{l}{LPIPS$\downarrow$}          & PSNR$\uparrow$                  & SSIM$\uparrow$                  & LPIPS$\downarrow$                     & PSNR$\uparrow$                      & SSIM$\uparrow$                                & \multicolumn{1}{l}{LPIPS$\downarrow$}          & PSNR$\uparrow$                  & SSIM$\uparrow$                  & LPIPS$\downarrow$          & PSNR$\uparrow$           & SSIM$\uparrow$           & GPU (ms)$\downarrow$ & \multicolumn{1}{l|}{Mobile (ms)$\downarrow$} & \multicolumn{1}{c}{GB}$\downarrow$ \\ \hline
AdaMPI \cite{han2022single} $\ddagger$ & -                                  & \multicolumn{1}{c}{-} & \multicolumn{1}{c}{-} & \multicolumn{1}{c}{0.303} & \multicolumn{1}{c}{20.46} & \multicolumn{1}{c|}{0.583}          & -                                  & \multicolumn{1}{c}{-} & \multicolumn{1}{c}{-} & 0.306          & 20.00          & 0.571          &    \multicolumn{1}{c}{254}      &  \multicolumn{1}{c|}{N/A}            &           0.15             \\
Ours                                                    & \multicolumn{1}{l}{\textbf{0.143}} & \textbf{26.87}        & \textbf{0.877}        & \textbf{ 0.088}            & \textbf{ 29.33}            & {\ul\hspace{0.1cm}0.883}                         & \multicolumn{1}{l}{\textbf{0.095}} & \textbf{26.87}        & \textbf{0.902}        & \textbf{0.069} & {\ul 29.03}    & {\ul 0.895}    &     \multicolumn{1}{c}{\textbf{26}}     & \multicolumn{1}{c|}{\textbf{33}}            &   \textbf{0.14 }                    \\ \hline
AdaMPI \cite{han2022single}            & -                                  & \multicolumn{1}{c}{-} & \multicolumn{1}{c}{-} & \multicolumn{1}{c}{0.100} & \multicolumn{1}{c}{26.23} & \multicolumn{1}{c|}{\textbf{0.888}} & -                                  & \multicolumn{1}{c}{-} & \multicolumn{1}{c}{-} & 0.078          & 26.38          & \textbf{0.909} & 254         & \multicolumn{1}{c|}{N/A}            &   0.15                     \\
Ours                                                    & \multicolumn{1}{l}{{\ul 0.158}}    & {\ul 25.90}           & {\ul 0.848}           & {\ul \hspace{0.15cm}0.099}               & {\ul  \hspace{0.15cm}28.57}               &  \hspace{0.1cm}0.865                               & \multicolumn{1}{l}{{\ul 0.100}}    & {\ul 26.82}           & {\ul 0.893}           & 0.071          & \textbf{29.31} & 0.892          &    \textbf{26}      & \multicolumn{1}{c|}{\textbf{33}}            &       \textbf{0.14}               \\ \hline
\end{tabular}}
\vspace{1mm}
    \caption{Results on COCO and Open Images. Rows 1 and 2 are models trained on Open Images, whereas the others are trained on COCO. OpenImages and COCO columns indicate evaluation on Open Images and COCO test sets. AdaMPI \cite{han2022single} performs the conventional warping we use as ground-truth, so we do not report its warping results. $\ddagger$ is AdaMPI trained on Open Images by us.} 
    \label{tab:comparison}
    \vspace{-7mm}
\end{table*}

\noindent \textbf{Flow decoder.} Having encoded the necessary inputs, we first focus on our flow decoder. The aim of this decoder is to take the shared embedding $\mathbf{F}$ as input and output the shift-map $\mathbb{S}$, which calculates the offset values to be applied on each pixel to perform the warping. Once the shift map is applied to the input image $\mathbb{I}_{s}$, we get the warped RGB image $\mathbb{I}_{warped}$, which is then multiplied by $\mathbb{M}$ to only contain the pixels coming from the source image. Note that our shift map network is conditioned on the target camera pose as the input $\mathbf{F}$ is conditioned on $\nabla(\cdot)$, unlike the stereo flow networks \cite{zhou2023single,chen2022pseudo,evain2019lightweight,chaurasia2020passthrough} that completely discard it.

\noindent \textbf{Mask decoder.}
The aim of the mask decoder is to generate the mask $\mathbb{M}$ necessary to perform the blending of Equation \ref{eq:cheapnvs}. Using the shared embedding $\mathbf{F}$ as input, the mask decoder predicts a binary mask that contains the spatial information to blend the warped RGB image and the result of inpainting. 

\noindent \textbf{Inpainting decoder.}
The aim of the inpainting decoder is to map the input features $\mathbf{F}$ to the dense inpainting output $\mathbb{P}$. This inpainting output is then multiplied with the inverse of $\mathbb{M}$, which filters out all the shifted pixels from the input image and adds only the inpainted colours to the final output.  

\noindent \textbf{Discussion.} These three decoders run concurrently and share the same architecture, which is formed of decoder blocks having bilinear upsampling, convolution and ELU activation steps. Conceptually, the warping is implemented by the flow and mask decoders, whereas the occlusion-filling is implemented by the inpainting decoder. Note that each of these decoders are working independently expect their shared encoders, which means they can also be used as isolated modules.
\vspace{-1mm}
\subsection{Multi-stage training}
\noindent Despite the parallelism-friendly formulation of Equation \ref{eq:cheapnvs}, in single-view NVS, inpainting is conceptually dependant on warping as the warping supplies inpainting inputs, namely the masked warped RGB and the mask itself. From a practical point of view, that necessitates a functional warping strategy before starting to optimize the inpainting module to specialise in narrow baseline occlusion masks. 

To this end, we devise a simple multi-stage training strategy, where we gradually activate new decoders in training. We first train the flow and mask decoders on their respective ground-truths, which helps the shared encoders and these two decoder supply rich information to the inpainting module. After a few epochs, we activate the training of the inpainting module, and train the entire pipeline until convergence.

\vspace{-0.5mm}






\section{Experimental Results}
\label{experiments}

\subsection{Experimental Details}
\noindent \textbf{Training.} We train our models with a combination of several losses. We use  L1 loss for the flow decoder, cross-entropy loss for the mask decoder. For the inpainting decoder we use L1, as we observe using the losses of \cite{han2022single} degrade our results. The overall loss is defined as 
\vspace{-3mm}

\begin{equation}
\begin{split}
    \mathcal{L}_{total} &= 
 \underbrace{\lambda_{1} \mathcal{L}_{L_1}}_{\mathcal{L}_{inpaint}} + \underbrace{\lambda_{2} \mathcal{L}_{CE}}_{\mathcal{L}_{mask}}+
 \underbrace{\lambda_{3} \mathcal{L}_{L_1}}_{\mathcal{L}_{flow}}
\end{split} \label{eq:losses}
\end{equation}

where at epoch 0, we set $\lambda_{1}=0$, $\lambda_{2}=\lambda_{3}=1$. Starting from the 5th epoch, we set $\lambda_{1}=\lambda_{2}=\lambda_{3}=1$. This setting facilitates our multi-stage training. We train our model for 20 epochs with a learning rate of $1e-4$ and a batch size of 32. We train on random crops of size 224x224 with horizontal flip augmentation, and use Adam optimizer \cite{kingma2014adam}.

\begin{figure*}[!ht]
  \centering
    \includegraphics[width=0.95\textwidth]{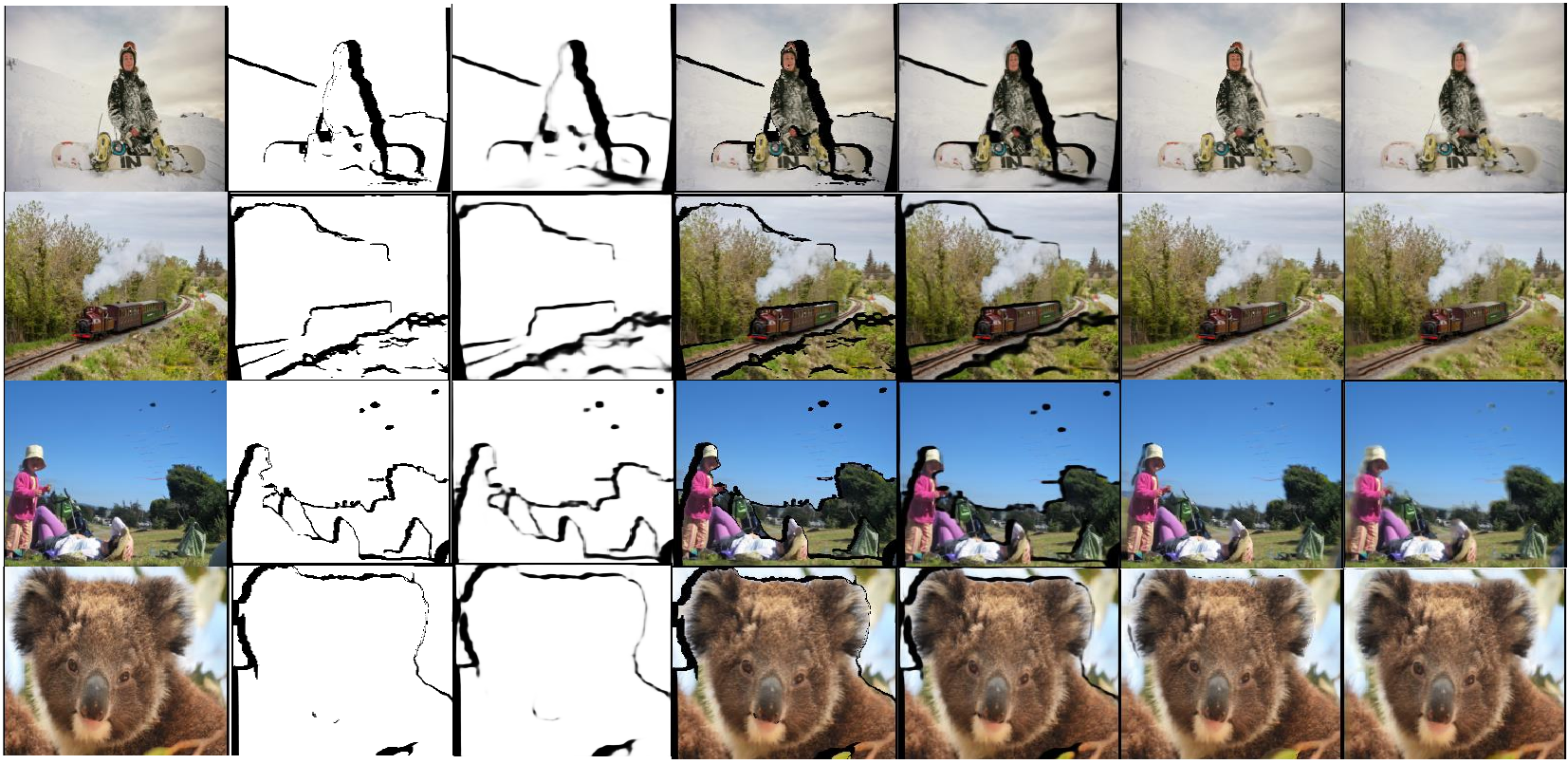}
  \vspace{-3mm}
  \caption{Left to right: Input, ground-truth occlusion mask, CheapNVS occlusion mask, ground-truth warped RGB, CheapNVS warped RGB, AdaMPI inpainting, CheapNVS inpainting. CheapNVS manages to handle object border artefacts in inpainting (1st, 3rd and 4th rows) and approximates warping successfully.}
  \label{fig:qualitative}
  \vspace{-6mm}
\end{figure*}

\noindent \textbf{Training datasets.} Following \cite{han2022single}, we use COCO to train and use 118K images for training. We also use Open Images \cite{kuznetsova2020open} as another training set, and randomly sample 174K images for training. We generate our ground-truths on-the-fly as \cite{han2022single}. For mask and flow decoders, we generate their labels through conventional 3D warping. We use the random transformation matrix generation approach of \cite{han2022single}, which effectively defines the narrow baseline definition throughout this paper. For inpainting, we create our own pseudo-labeler; we finetune a pretrained MI-GAN \cite{sargsyan2023mi} on warp-back data using OpenImages. This model is then used to generate inpainted images which are used as ground-truth for the inpainting decoder. 

\noindent \textbf{Evaluation datasets.} In ablations, we use OpenImages to train and evaluate, and randomly select 12.5K images for testing. For comparison with state-of-the-art, we use the same randomly selected 12.5K images from OpenImages and the test set of COCO. We generate the ground-truths the same way as we do in training.

\begin{table}[!t]
\resizebox{0.48\textwidth}{!}{%
\begin{tabular}{lllllll}
\hline
& \multicolumn{3}{c}{Warping}                      & \multicolumn{3}{c}{Inpainting}                   \\
& LPIPS$\downarrow$          & PSNR$\uparrow$           & SSIM$\uparrow$           & LPIPS$\downarrow$          & PSNR$\uparrow$           & SSIM$\uparrow$           \\ \hline
$\mathcal{L}_{inpaint}$        & 0.211          & 22.43          & 0.778          & 0.095          & \textbf{29.56} & 0.877          \\
$\mathcal{L}_{inpaint}$ + $\mathcal{L}_{mask}$  & {\ul 0.147}    & {\ul 26.55}    & {\ul 0.864}    & {\ul 0.090}    & 29.31          & {\ul 0.878} \\
$\mathcal{L}_{total}$              & \textbf{0.143} & \textbf{26.87} & \textbf{0.877} & \textbf{0.088} & {\ul 29.33}    & \textbf{0.883} \\ \hline
No SC                 & 0.198          & 24.91          & {\ul 0.820}    & 0.109          & {\ul 28.64}    & {\ul 0.869} \\
SC to all decoders    & {\ul 0.153}    & {\ul 25.93}    & {\ul 0.820}    & {\ul 0.102}    & 28.10          & 0.828 \\
SC to $f(\cdot)$ and  $\phi(\cdot)$ & \textbf{0.143} & \textbf{26.87} & \textbf{0.877} & \textbf{0.088} & \textbf{29.33} & \textbf{0.883} \\ \hline
L1 + FFL + Perc + SSIM & 0.154 & 25.73 & 0.827 & 0.093 & 28.54 & 0.853 \\ 
L1 + SSIM  & 0.152 & 26.56 & 0.866 & 0.091 & {\ul 29.32} & \textbf{0.888} \\ 
L1 + Perc  & \textbf{0.142} & {\ul 26.69} & {\ul 0.872} & {\ul 0.089} & 29.21 & 0.879 \\ 
L1 + FFL  & 0.147 & 26.46 & 0.858 & 0.093 & 28.54 & 0.853 \\ 
L1 & {\ul 0.143} & \textbf{26.87} & \textbf{0.877} & \textbf{0.088} & \textbf{29.33} & {\ul 0.883} \\ \hline
\end{tabular}%
}

\begin{minipage}{0.98\columnwidth}
  \caption{Ablation results on OpenImages test set. \textbf{SC} stands for skip connections from RGBD encoder. \textbf{FFL} and \textbf{Perc} indicate the usage of focal frequency \cite{jiang2021focal} and perceptual \cite{zhang2018unreasonable} losses.}
  \label{tab:ablation}
\end{minipage}
\vspace{-7.5mm}
\end{table}


\noindent \textbf{Metrics.} We evaluate CheapNVS in warping as well as inpainting, and use SSIM, PSNR and LPIPS \cite{zhang2018unreasonable} metrics. For warping, the ground-truth warping is the conventional depth-based 3D warping implemented in PyTorch. We use DPT \cite{ranftl2021vision} depth maps for COCO training to be fair against AdaMPI, and Marigold \cite{ke2024repurposing} depth maps for Open Images training.

\subsection{Results}
\noindent We compare CheapNVS against AdaMPI \cite{han2022single}, the current state-of-the-art method for 3D photography. We use its original weights, but we also train it using the original code on OpenImages for a fair comparison against our method. We omit other older methods from our comparison, as they are already inferior to AdaMPI in accuracy.

\noindent \textbf{Quantitative.} The results shown in Table \ref{tab:comparison} indicate that CheapNVS, except SSIM scores, is in fact better than AdaMPI in inpainting on both Open Images and COCO test sets. This shows that our lightweight decoder manages to do as well as the heavy multi-decoder approach of AdaMPI. We credit this to CheapNVS' shared latent space between multiple decoders that facilitate rich interaction between them. As AdaMPI uses the conventional 3D warping pipeline that we use to generate our ground-truths, the warping results in Table \ref{tab:comparison} shows how far off we are from the ground-truth. The results show that we are doing fine in approximating the laborious 3D warping with our cheap double encoders. The inpainting results also verify that, as inpainting inherently relies on the warping accuracy. 

Finally, our model trained on Open Images beat our model trained on COCO and the original AdaMPI trained on COCO. We also train AdaMPI on Open Images with its original settings, but this training fails to provide a tangible result. Ultimately, the results show that Open Images as a training dataset can offer better scale, and manages to outperform COCO as a training set for narrow baseline NVS.

\noindent \textbf{Qualitative.} We provide visual results in Figure \ref{fig:qualitative}. CheapNVS manages to perform warping and occlusion mask prediction successfully. Specifically, it not only approximates the ground-truth predictions quite well, but manages to \textit{smooth} the occlusion masks by itself (see first and third rows of Figure \ref{fig:qualitative}). Note that occlusion masks are often post-processed with operations such as smoothing, connected components and dilation/erosion in conventional methods, as these are necessary to perform an accurate blending. CheapNVS learns to do these automatically, as it jointly learns inpainting as well. 

In inpainting, against  much more complex competitors, CheapNVS still performs competitively. CheapNVS actually does a better job in object boundary artefact removal compared to AdaMPI (see the person on 1st and 3rd row, and Koala on the last row). In cases where there are large translations on the image boundaries, such as left edge of image on the second row, AdaMPI exhibits stretching artefacts, whereas CheapNVS avoids this behaviour.

\noindent \textbf{Runtime.} Table \ref{tab:comparison} shows that CheapNVS is worthy of its name and runs at 26 ms, which is 10 $\times$ faster than AdaMPI on an RTX 3090 GPU. Furthermore, CheapNVS has lower memory consumption during inference (a batch size of 1). CheapNVS runs at 33ms (via TFLite GPU delegate) on a Samsung Tab 9+, whereas AdaMPI can not be ported naively as it uses external libraries for 3D warping. Note that porting the conventional image warping efficiently would require domain expertise, yet CheapNVS can be seamlessly run on various hardware accelerators without requiring such expertise.  Other methods \cite{kopf2020one} report 800ms on an IPhone with resolution 1152x1536 or 300ms on a desktop P100 GPU with 762x1008 resolution \cite{jampani2021slide}. Once lifted to these resolutions, CheapNVS takes 618 and 275ms,  which is still much faster than \cite{jampani2021slide,kopf2020one}.

\vspace{-1mm}
\subsection{Ablations} \label{sec:ablations}
\noindent \textbf{Supervision.} Our method leverages three distinct supervisory losses, one for each decoder, as shown in Equation \ref{eq:losses}. The first three rows of Table \ref{tab:ablation} shows that addition of each loss term helps the final warping and inpainting accuracy. Furthermore, the first row shows the possibility of learning warping via purely inpainting supervision, paving the way for weak supervised methods for learnable warping. 

\noindent \textbf{Skip connections.} As Figure \ref{fig:overall_diagram} shows, we have skip connections from RGBD encoder to inpainting and mask decoders. The second part (rows 3 to 6) of Table \ref{tab:ablation} shows the effect of adding such skip connections. Adding skip connections definitely help the final results both for warping and inpainting. Adding a skip connection to flow decoder does the opposite and degrades the results. We believe the reason is the nature of each decoders' output; inpainting and mask decoders predict outputs that are spatially-meaningful and relevant to the content of the input image. The flow decoder, however, predicts a shift-map that has no relation with the other two decoders' output. This discrepancy, we believe, is the reason why skip connections to the flow decoder harm the results.

\noindent \textbf{Losses.} AdaMPI uses L1, focal frequency \cite{jiang2021focal}, perceptual \cite{zhang2018unreasonable} and SSIM losses for training. We explore the loss design space, and observe that we obtain better quantitative results by using only the L1 loss. As shown in the last rows of Table \ref{tab:ablation}, L1 loss training obtains best results in warping and inpainting, with close second being L1 + perceptual loss. We do not observe tangible qualitative differences between the results of training with or without the perceptual loss.
\vspace{-0.5mm}
\subsection{Limitations and Future Work}

\noindent A common failure mode of CheapNVS is in warping, which ultimately effects the quality of the synthesized view. Similarly, since we learn the warping on a narrow baseline specified during the training, CheapNVS might struggle to warp in baselines that are larger than what it is trained on, whereas AdaMPI might do well since it performs the highly expensive, handcrafted warping. We believe this to be the reason why we outperform AdaMPI in narrow baselines despite its use of complex multi-plane images and multiple inpainting networks, whereas they might do better than us in larger baselines (such as some stereo datasets). We note, however, that these larger baselines are not within the scope of our work. In future work, we aim to alleviate this warping domain gap, and also improve synthesis quality by using a more powerful inpainting teacher.

\vspace{-0.5mm}

\section{Conclusion}
\label{conclusion}
\noindent We propose CheapNVS, a narrow-baseline single-view novel view synthesis method that comfortably runs on real-time on mobile devices. CheapNVS approximates the laborious depth-based 3D image warping by efficient modules that learn to perform warping conditioned on the target view. Furthermore, we cast single-view NVS as an image blending problem, and perform warping and inpainting concurrently to save computational resources. Trained in multiple stages and on a diverse, large-scale dataset, CheapNVS outperforms the existing state-of-the-art method while performing up to 10 $\times$ faster and consuming 6\% less memory.


\bibliographystyle{IEEEbib}
\bibliography{refs}

\begin{thebibliography}{10}

\bibitem{adamkiewicz2022vision}
Michal Adamkiewicz, Timothy Chen, Adam Caccavale, Rachel Gardner, Preston Culbertson, Jeannette Bohg, and Mac Schwager,
\newblock ``Vision-only robot navigation in a neural radiance world,''
\newblock {\em IEEE Robotics and Automation Letters}, vol. 7, no. 2, pp. 4606--4613, 2022.

\bibitem{li2023instant}
Sixu Li, Chaojian Li, Wenbo Zhu, Boyang Yu, Yang Zhao, Cheng Wan, Haoran You, Huihong Shi, and Yingyan Lin,
\newblock ``Instant-3d: Instant neural radiance field training towards on-device ar/vr 3d reconstruction,''
\newblock in {\em Proceedings of the 50th Annual International Symposium on Computer Architecture}, 2023, pp. 1--13.

\bibitem{skartados2024finding}
Evangelos Skartados, Mehmet~Kerim Yucel, Bruno Manganelli, Anastasios Drosou, and Albert Sa{\`a}-Garriga,
\newblock ``Finding waldo: Towards efficient exploration of nerf scene spaces,''
\newblock in {\em Proceedings of the 15th ACM Multimedia Systems Conference}, 2024, pp. 155--165.

\bibitem{corona2022mednerf}
Abril Corona-Figueroa, Jonathan Frawley, Sam Bond-Taylor, Sarath Bethapudi, Hubert~PH Shum, and Chris~G Willcocks,
\newblock ``Mednerf: Medical neural radiance fields for reconstructing 3d-aware ct-projections from a single x-ray,''
\newblock in {\em 2022 44th annual international conference of the IEEE engineering in medicine \& Biology society (EMBC)}. IEEE, 2022, pp. 3843--3848.

\bibitem{mildenhall2021nerf}
Ben Mildenhall, Pratul~P Srinivasan, Matthew Tancik, Jonathan~T Barron, Ravi Ramamoorthi, and Ren Ng,
\newblock ``Nerf: Representing scenes as neural radiance fields for view synthesis,''
\newblock {\em Communications of the ACM}, vol. 65, no. 1, pp. 99--106, 2021.

\bibitem{kerbl20233d}
Bernhard Kerbl, Georgios Kopanas, Thomas Leimk{\"u}hler, and George Drettakis,
\newblock ``3d gaussian splatting for real-time radiance field rendering.,''
\newblock {\em ACM Trans. Graph.}, vol. 42, no. 4, pp. 139--1, 2023.

\bibitem{fan2023pose}
Zhiwen Fan, Panwang Pan, Peihao Wang, Yifan Jiang, Hanwen Jiang, Dejia Xu, Zehao Zhu, Dilin Wang, and Zhangyang Wang,
\newblock ``Pose-free generalizable rendering transformer,''
\newblock {\em arXiv e-prints}, pp. arXiv--2310, 2023.

\bibitem{zhou2023single}
Yang Zhou, Hanjie Wu, Wenxi Liu, Zheng Xiong, Jing Qin, and Shengfeng He,
\newblock ``Single-view view synthesis with self-rectified pseudo-stereo,''
\newblock {\em International Journal of Computer Vision}, vol. 131, no. 8, pp. 2032--2043, 2023.

\bibitem{chen2022pseudo}
Yi-Nan Chen, Hang Dai, and Yong Ding,
\newblock ``Pseudo-stereo for monocular 3d object detection in autonomous driving,''
\newblock in {\em Proceedings of the IEEE/CVF conference on computer vision and pattern recognition}, 2022, pp. 887--897.

\bibitem{evain2019lightweight}
Simon Evain and Christine Guillemot,
\newblock ``A lightweight neural network for monocular view generation with occlusion handling,''
\newblock {\em IEEE Transactions on Pattern Analysis and Machine Intelligence}, vol. 43, no. 6, pp. 1832--1844, 2019.

\bibitem{chaurasia2020passthrough}
Gaurav Chaurasia, Arthur Nieuwoudt, Alexandru-Eugen Ichim, Richard Szeliski, and Alexander Sorkine-Hornung,
\newblock ``Passthrough+ real-time stereoscopic view synthesis for mobile mixed reality,''
\newblock {\em Proceedings of the ACM on Computer Graphics and Interactive Techniques}, vol. 3, no. 1, pp. 1--17, 2020.

\bibitem{shih20203d}
Meng-Li Shih, Shih-Yang Su, Johannes Kopf, and Jia-Bin Huang,
\newblock ``3d photography using context-aware layered depth inpainting,''
\newblock in {\em Proceedings of the IEEE/CVF Conference on Computer Vision and Pattern Recognition}, 2020, pp. 8028--8038.

\bibitem{han2022single}
Yuxuan Han, Ruicheng Wang, and Jiaolong Yang,
\newblock ``Single-view view synthesis in the wild with learned adaptive multiplane images,''
\newblock in {\em ACM SIGGRAPH 2022 Conference Proceedings}, 2022, pp. 1--8.

\bibitem{shi2023zero123}
Ruoxi Shi, Hansheng Chen, Zhuoyang Zhang, Minghua Liu, Chao Xu, Xinyue Wei, Linghao Chen, Chong Zeng, and Hao Su,
\newblock ``Zero123++: a single image to consistent multi-view diffusion base model,''
\newblock {\em arXiv preprint arXiv:2310.15110}, 2023.

\bibitem{tang2024mvdiffusion}
Shitao Tang, Jiacheng Chen, Dilin Wang, Chengzhou Tang, Fuyang Zhang, Yuchen Fan, Vikas Chandra, Yasutaka Furukawa, and Rakesh Ranjan,
\newblock ``Mvdiffusion++: A dense high-resolution multi-view diffusion model for single or sparse-view 3d object reconstruction,''
\newblock {\em arXiv preprint arXiv:2402.12712}, 2024.

\bibitem{kopf2020one}
Johannes Kopf, Kevin Matzen, Suhib Alsisan, Ocean Quigley, Francis Ge, Yangming Chong, Josh Patterson, Jan-Michael Frahm, Shu Wu, Matthew Yu, et~al.,
\newblock ``One shot 3d photography,''
\newblock {\em ACM Transactions on Graphics (TOG)}, vol. 39, no. 4, pp. 76--1, 2020.

\bibitem{schoenberger2016sfm}
Johannes~Lutz Sch\"{o}nberger and Jan-Michael Frahm,
\newblock ``Structure-from-motion revisited,''
\newblock in {\em Conference on Computer Vision and Pattern Recognition (CVPR)}, 2016.

\bibitem{jantet2009incremental}
Vincent Jantet, Luce Morin, and Christine Guillemot,
\newblock ``Incremental-ldi for multi-view coding,''
\newblock in {\em 2009 3DTV Conference: The True Vision-Capture, Transmission and Display of 3D Video}. IEEE, 2009, pp. 1--4.

\bibitem{flynn2019deepview}
John Flynn, Michael Broxton, Paul Debevec, Matthew DuVall, Graham Fyffe, Ryan Overbeck, Noah Snavely, and Richard Tucker,
\newblock ``Deepview: View synthesis with learned gradient descent,''
\newblock in {\em Proceedings of the IEEE/CVF Conference on Computer Vision and Pattern Recognition}, 2019, pp. 2367--2376.

\bibitem{georgiadis2022adaptive}
Konstantinos Georgiadis, Albert Sa{\`a}-Garriga, Mehmet~Kerim Yucel, Anastasios Drosou, and Bruno Manganelli,
\newblock ``Adaptive mask-based pyramid network for realistic bokeh rendering,''
\newblock in {\em European Conference on Computer Vision}. Springer, 2022, pp. 429--444.

\bibitem{yucel2023lra}
Mehmet~Kerim Y{\"u}cel, Valia Dimaridou, Bruno Manganelli, Mete Ozay, Anastasios Drosou, and Albert Saa-Garriga,
\newblock ``Lra\&ldra: Rethinking residual predictions for efficient shadow detection and removal,''
\newblock in {\em Proceedings of the IEEE/CVF Winter Conference on Applications of Computer Vision}, 2023, pp. 4925--4935.

\bibitem{sandler2018mobilenetv2}
Mark Sandler, Andrew Howard, Menglong Zhu, Andrey Zhmoginov, and Liang-Chieh Chen,
\newblock ``Mobilenetv2: Inverted residuals and linear bottlenecks,''
\newblock in {\em Proceedings of the IEEE conference on computer vision and pattern recognition}, 2018, pp. 4510--4520.

\bibitem{kingma2014adam}
Diederik Kingma and Jimmy Ba,
\newblock ``Adam: A method for stochastic optimization,''
\newblock in {\em International Conference on Learning Representations (ICLR)}, San Diega, CA, USA, 2015.

\bibitem{kuznetsova2020open}
Alina Kuznetsova, Hassan Rom, Neil Alldrin, Jasper Uijlings, Ivan Krasin, Jordi Pont-Tuset, Shahab Kamali, Stefan Popov, Matteo Malloci, Alexander Kolesnikov, et~al.,
\newblock ``The open images dataset v4: Unified image classification, object detection, and visual relationship detection at scale,''
\newblock {\em International journal of computer vision}, vol. 128, no. 7, pp. 1956--1981, 2020.

\bibitem{sargsyan2023mi}
Andranik Sargsyan, Shant Navasardyan, Xingqian Xu, and Humphrey Shi,
\newblock ``Mi-gan: A simple baseline for image inpainting on mobile devices,''
\newblock in {\em Proceedings of the IEEE/CVF International Conference on Computer Vision}, 2023, pp. 7335--7345.

\bibitem{jiang2021focal}
Liming Jiang, Bo~Dai, Wayne Wu, and Chen~Change Loy,
\newblock ``Focal frequency loss for image reconstruction and synthesis,''
\newblock in {\em Proceedings of the IEEE/CVF international conference on computer vision}, 2021, pp. 13919--13929.

\bibitem{zhang2018unreasonable}
Richard Zhang, Phillip Isola, Alexei~A Efros, Eli Shechtman, and Oliver Wang,
\newblock ``The unreasonable effectiveness of deep features as a perceptual metric,''
\newblock in {\em Proceedings of the IEEE conference on computer vision and pattern recognition}, 2018, pp. 586--595.

\bibitem{ranftl2021vision}
Ren{\'e} Ranftl, Alexey Bochkovskiy, and Vladlen Koltun,
\newblock ``Vision transformers for dense prediction,''
\newblock in {\em Proceedings of the IEEE/CVF international conference on computer vision}, 2021, pp. 12179--12188.

\bibitem{ke2024repurposing}
Bingxin Ke, Anton Obukhov, Shengyu Huang, Nando Metzger, Rodrigo~Caye Daudt, and Konrad Schindler,
\newblock ``Repurposing diffusion-based image generators for monocular depth estimation,''
\newblock in {\em Proceedings of the IEEE/CVF Conference on Computer Vision and Pattern Recognition}, 2024, pp. 9492--9502.

\bibitem{jampani2021slide}
Varun Jampani, Huiwen Chang, Kyle Sargent, Abhishek Kar, Richard Tucker, Michael Krainin, Dominik Kaeser, William~T Freeman, David Salesin, Brian Curless, et~al.,
\newblock ``Slide: Single image 3d photography with soft layering and depth-aware inpainting,''
\newblock in {\em Proceedings of the IEEE/CVF International Conference on Computer Vision}, 2021, pp. 12518--12527.

\end{thebibliography}
\end{document}